\documentclass[conference]{IEEEtran}
\IEEEoverridecommandlockouts

\usepackage{cite}
\usepackage{amsmath,amssymb,amsfonts}
\usepackage{algorithmic}
\usepackage{graphicx}
\usepackage{textcomp}
\usepackage{xcolor}
\def\BibTeX{{\rm B\kern-.05em{\sc i\kern-.025em b}\kern-.08em
    T\kern-.1667em\lower.7ex\hbox{E}\kern-.125emX}}

\begin{document}

\title{Systematic review of image segmentation using complex networks}

\author{\IEEEauthorblockN{1\textsuperscript{st} Amin Rezaei}
    \IEEEauthorblockA{
        \textit{IEEE Member} \\
        Yazd, Iran \\
        amin.rezaei@ieee.org
    }
\and
\IEEEauthorblockN{2\textsuperscript{nd} Fateme Asadi}
    \IEEEauthorblockA{
        \textit{Yazd University} \\
        Fars, Iran \\
        fatemeh.asadi@stu.yazd.ac.ir
    }
}

\maketitle

\begin{abstract}
    This review presents various image segmentation methods using complex networks. 
    Image segmentation is one of the important steps in image analysis as it helps analyze and understand complex images. At first, it has been tried to classify complex networks based on how it being used in image segmentation.
    In computer vision and image processing applications, image segmentation is essential for analyzing complex images with irregular shapes, textures, or overlapping boundaries. Advanced algorithms make use of machine learning, clustering, edge detection, and region-growing techniques. Graph theory principles combined with community detection-based methods allow for more precise analysis and interpretation of complex images. Hybrid approaches combine multiple techniques for comprehensive, robust segmentation, improving results in computer vision and image processing tasks.
\end{abstract}


\section{INTRODUCTION}

    Image segmentation is essential in computer vision and image processing applications, as it helps analyze and understand complex images with irregular shapes, textures, or overlapping boundaries. Advanced image segmentation algorithms are required to accurately identify and separate different objects or regions within an image. These algorithms utilize various techniques such as edge detection, region growing, clustering, and machine learning. 
    
    Edge detection is a common technique used in image segmentation to identify boundaries between different objects or regions. On the other hand, region-growing algorithms separate pixels with comparable characteristics into distinct regions. Additionally, pixels can be grouped based on their similarity in color or intensity using clustering techniques like k-means clustering. Furthermore, machine learning approaches, such as convolutional neural networks (CNNs), have shown promising results in automatically segmenting images by learning from large datasets. However, traditional segmentation techniques often struggle with complex images, but community detection-based approaches have emerged as a promising solution. These methods use graph theory principles to identify groups of pixels with strong intra-group similarities while maintaining weak inter-group connections. By utilizing community detection-based approaches, segmentation algorithms can effectively identify regions within an image that share similar characteristics. This enables more accurate analysis and interpretation of complex images, ultimately enhancing the performance of computer vision and image processing applications. 
    
    Additionally, there are also hybrid approaches that combine multiple segmentation techniques to overcome the limitations of individual methods. These hybrid approaches can provide a more comprehensive and robust segmentation of images by leveraging the strengths of different algorithms. By combining the outputs of multiple techniques, they can overcome the weaknesses and limitations that may arise from using a single segmentation method. This allows for a more accurate and detailed analysis of complex images, leading to improved results in computer vision and image processing tasks.

\section{LITERATURE REVIEW}

    A hybrid approach of dynamic image processing and complex network proposed in \cite{ebrahimi2023hybrid}, which identifies repetitive images of welding defects in radiographs of oil and gas pipelines. The study of welding in gas and oil pipelines is crucial for non-destructive testing, and expert interpreters are used to interpret radiographic films. The area growth method is used to segment images and identify welding defects, but it has limitations in images with less subject variety. The proposed method uses a histogram to determine the start and end images of the welding range, then a combination of standard algorithms is applied to identify defects. Key points of the image are extracted, the corresponding complex dynamic network is drawn and its calculations are performed.
    
    \cite{gammoudi2021brain} presents a novel approach for image segmentation based on community detection algorithms existing in social networks. The authors propose a method based on super-pixels and algorithms for community detection in graphs. The super-pixel method reduces the number of nodes in the graph, while community detection algorithms provide more accurate segmentation than traditional approaches. The method is compared with the image segmentation method based on deep learning and previous work. Experimental results show that the method provides more precise segmentation.
    
    An image retrieval system based on a complex network model proposed in \cite{moghadam2021multilayered}. A multilayered complex network is constructed between images of each category based on color, texture, and shape features. A meta-path is defined as the way of connecting two images in the network, and a set of informative meta-paths are composed to find similar images by exploring the network. The established complex network provides an efficient way to benefit from image correlations to enhance the similarity search of the images. Employing diverse meta-paths with different semantics leads to measuring image similarities based on effective image features for each category.
    
    A new approach to interactive image segmentation using Particle Competition and Cooperation (PCC) model was proposed in \cite{passerini2020complex}, which eliminates the weight vector in the network construction phase. The proposed model and the reference model, without the use of a weight vector, were compared using 151 images extracted from the Grabcut dataset, PASCAL VOC dataset, and Alpha matting dataset. The simulations resulted in an error rate of only 0.49\% when classifying pixels with the proposed model, while the reference model had an error rate of 3.14\%. The proposed method also presented less error variation in the diversity of the evaluated images compared to the reference model. The paper highlights the importance of using user input as clues to the algorithm in image segmentation, as it is considered one of the hardest tasks in image processing.
    
    A simple community detection framework proposed in \cite{freitas2019community}, covers the graph construction process from feature vectors generated from non-graph data until the application and evaluation of community detection methods over such a graph. The framework is further evaluated on the problem of invariant pattern clustering of images, which consists of giving a set of image objects taken from different angles, positions, or rotations, clustering the images related to each object.
    
    \cite{mourchid2019general} proposes an image segmentation general framework using complex networks-based community detection algorithms to segment images into meaningful connected components. The authors start by splitting the image into small regions using initial segmentation, which is then used to build the complex network. To produce meaningful connected components and detect homogeneous communities, color, and texture-based features are employed to quantify region similarities. The network of regions is constructed adaptively to avoid many small regions in the image, and community detection algorithms are applied to the resultant adaptive similarity matrix to obtain the final segmented image.
    
    The research conducted in \cite{lai2019material} focuses on the study of the relationship between manufacturing processes, material structure, and properties of materials to develop new materials. Material images contain microstructures of materials, and quantitative analysis is crucial for studying their characteristics. However, most material microstructures are shown with various shapes and complex textures, which hinder the exact segmentation of component elements. The study adopts machine learning and complex network methods to address this challenge. Two segmentation tasks are completed: one for titanium alloy images, which are segmented based on pixel-level classification through feature extraction and machine learning algorithms, and another for ceramic images, which are segmented using complex networks theory.Why do Fortune 500 companies and top research institutions trust Overleaf to streamline their collaboration? Get in touch to learn more.
    
    \cite{mourchid2017image} proposed a new perspective on image segmentation by applying two efficient community detection algorithms. By considering regions as communities, these methods can give an over-segmented image of many small regions. The proposed algorithms are improved to automatically merge neighboring regions additively to achieve the highest modularity/stability. To produce sizable regions and detect homogeneous communities, the authors use the combination of a feature based on the Histogram of Oriented Gradients of the image and a feature based on color to characterize the similarity between the two regions. By constructing the similarity matrix in an adaptive manner, they avoid the problem of over-segmentation.
    
    \cite{trufanov2017image} proposed scale segmentation views, local, medium, and global for an image to build a genuine complex network. The case study with two sample images demonstrates how the scales are connected with the formation of network topology. The role of network topology is of great importance as it stipulates functional specificities inherent to the system.
    
    Focusing on segmenting input images using efficient community detection algorithms from social and complex networks, \cite{abin2014wisecode} proposed a method that fragments the input image into small initial regions, then constructs a weighted network with mapped vertex connections. The similarity between two regions is calculated from color information, which is then used to assign weights to edges. A community detection algorithm is applied, extracting communities with the highest modularity measure. A post-processing algorithm merges small regions with larger ones, enhancing the final result.

\section{OVERVIEW OF IMAGE SEGMENTATION TECHNIQUES}

    \subsection*{Thresholding-based Segmentation}

        In this method, the intended region is defined as any pixels that are either above or below the threshold value. It is a straightforward but efficient method for separating images with different intensities.

    \subsection*{Clustering-based Segmentation}

        This method groups pixels depending on their color, texture, or other characteristics. k-means and hierarchical clustering are two popular clustering algorithms.

    \subsection*{Region-based Segmentation}
    
        This approach views the image as a collection of regions, or superpixels, where each region represents a coherent image area. Techniques such as graph-based segmentation and watershed transform are frequently used in this approach.

    \subsection*{Edge-based Segmentation}

        this method focuses on detecting boundaries or edges between different regions in the image. Techniques such as Canny edge detection and gradient-based methods are popular in this category.

    \subsection*{Comparison of Different Segmentation Techniques}

        Each segmentation method has benefits and drawbacks. The technique selected will depend on the unique image properties and the intended segmentation result. The best method for a problem can be chosen through comparison and experimentation.

\section{COMPLEX NETWORKS-BASED IMAGE SEGMENTATION METHODS AND APPROACHES}

    \subsection*{Traditional Community Detection Techniques}
    
        Algorithms like modularity optimization, Girvan-Newman, and Louvain, which concentrate on optimizing the modularity score or recognizing communities based on network topology, are examples of traditional community detection strategies.

    \subsection*{Recent Advancements in Community Detection Algorithms}

        Community detection algorithms have evolved to meet the demands of modern data analysis. Recent advancements include algorithms based on spectral clustering, deep learning, and machine learning approaches. These methods offer improved accuracy and efficiency in detecting communities within complex networks.

    \subsection*{Modularity Optimization for Image Segmentation}
    
        Modularity optimization is an effective approach for segmenting images. The goal is to optimize modularity, which assesses how well these parts are connected while minimizing connections between various segments. Techniques like spectral clustering or graph-based algorithms, effectively take advantage of the underlying structure within an image. By iteratively changing region bounds based on modularity scores, this method can produce very accurate segmentations. It allows for flexibility in setting the number of output segments and is suitable for various types of images, including irregular shapes and textures.

    \subsection*{Spectral Clustering for Community Detection in Images}

        Spectral clustering is a powerful method for detecting hidden patterns and structures in large image datasets. It uses graph theory concepts to represent images as graphs, with edges defined based on pixel similarity. This technique has proven successful in applications like object recognition, scene understanding, and social network analysis. Spectral clustering captures non-linear relationships effectively, enabling the identification of complex communities that may not be apparent using traditional methods. It outperforms other algorithms in terms of accuracy and efficiency when dealing with images containing multiple interconnected regions or overlapping communities.

    \subsection*{Graph-Cut based Segmentation using Community Detection}

        This method is an advanced image processing technique that accurately segments images into regions or objects of interest. It combines graph cuts and community detection algorithms to achieve this task. Graph cuts define energy functions that measure pixel dissimilarity, allowing for efficient partitioning of images into meaningful segments. Community detection algorithms identify clusters with strong interconnections, indicating potential regions or objects.

    \subsection*{Deep Learning Approaches for Community Detection-Based Segmentation}

        Deep learning approaches use advanced algorithms and neural network architectures to extract meaningful patterns and identify distinct communities within complex networks. These methods automatically discover hidden structures and relationships, enabling efficient segmentation of large-scale networks. Deep neural networks enable models to learn high-level representations, capturing complex features and characteristics of the network graph. These approaches can handle dynamic networks by adapting over time, facilitating the identification of evolving communities.

\section{APPLICATIONS AND CASE STUDIES}

    Medical image segmentation using community detection is a cutting-edge technique that accurately identifies and delineates anatomical structures in medical images. This emerging field combines image analysis, machine learning, and graph theory to improve diagnostic accuracy and treatment planning in healthcare. By partitioning the image into distinct regions, community detection algorithms like spectral clustering or graph-cuts efficiently segment the medical image. This technique aids in early disease diagnosis and monitoring therapy response. Although still in its experimental stage, medical image segmentation using community detection shows great promise for enhancing clinical workflows and patient outcomes in various medical domains, including radiology, pathology, and neurology.

    \subsection*{Semantic Segmentation in Natural Images}
    
        Semantic segmentation, which aims to assign meaningful labels to each pixel in an image, is another area where community detection-based methods have made significant contributions. By identifying communities or regions within an image that share similar semantic attributes, these algorithms enable a high-level understanding and interpretation of visual scenes. Applications of semantic segmentation include autonomous driving, image annotation, and scene understanding.

    \subsection*{Object Detection and Recognition using Community Detection}
    
        Detection and recognition of objects based on Community detection is a cutting-edge computer vision technique for identifying and classifying objects in images or videos. It employs algorithms to identify communities or groups of similar objects based on visual characteristics such as shape, color, texture, and size. This method improves the accuracy and efficiency of object detection models by reducing the search space for potential objects. It also allows for object recognition and labeling within each detected community, allowing for tasks such as scene comprehension, action recognition, and tracking.

\section{EVALUATION METRICS AND PERFORMANCE ANALYSIS}

    \subsection*{Quantitative Measures for Image Segmentation Evaluation}

        To evaluate image segmentation algorithms, quantifiable measures are essential for understanding their performance and comparing techniques. Common metrics include precision, recall, F1-score, and Intersection over Union (IoU), providing insights into accuracy, completeness, and effectiveness of segmentation results.

    \subsection*{Comparative Analysis of Community Detection-Based Segmentation Methods}
    
        A comparative analysis of community detection-based segmentation methods is crucial for understanding the strengths and limitations of different approaches. Community detection algorithms like Louvain, Leiden, and Infomap are popular due to their ability to discover meaningful clusters in complex networks. This analysis evaluates these algorithms based on accuracy, robustness, scalability, and computational efficiency, allowing researchers to choose the most suitable algorithm for their specific needs and network characteristics.

    \subsection*{Challenges and Limitations in Existing Evaluation Metrics}
    
        While evaluation metrics play a crucial role in assessing the performance of image segmentation algorithms, they are not without their limitations. One such challenge is the difficulty in defining ground truth annotations for complex images with intricate boundaries and overlapping objects. Additionally, existing metrics may not fully capture the perceptual quality of the segmentation results, as they primarily focus on pixel-wise accuracy. This limitation calls for the development of new evaluation metrics that can better reflect the human perception of segmentation quality.

\section{CHALLENGES AND FUTURE DIRECTIONS}

    \subsection*{Addressing Over-segmentation and Under-segmentation Issues}
    
        One of the major challenges faced by community detection-based image segmentation algorithms is the issue of over-segmentation or under-segmentation. Over-segmentation occurs when an image is divided into too many small regions, making it difficult to distinguish individual objects. Under-segmentation, on the other hand, merges multiple objects into a single region, leading to inaccurate segmentation. Future research should focus on developing algorithms that can address these challenges and achieve more precise and meaningful segmentation.

    \subsection*{Incorporating Contextual Information for Improved Segmentation}
    
        Contextual information, such as object relationships and scene context, can improve the accuracy and robustness of community detection-based segmentation significantly. Algorithms can better capture the boundaries and structures within an image by taking into account the interactions between objects and their contextual information. Future research should explore methods to effectively incorporate contextual information into the segmentation process, to achieve better results in complex scenarios.

    \subsection*{Integration of Deep Learning and Community Detection for Fine-grained Segmentation}
    
        Deep learning techniques have revolutionized image processing tasks, including image segmentation. The integration of deep learning and community detection holds great potential for achieving fine-grained segmentation, where objects with subtle differences can be accurately distinguished. By leveraging the power of deep neural networks to learn discriminative features, combined with the ability of community detection to capture structural relationships, this hybrid approach can push the boundaries of image segmentation performance even further. Future research should focus on developing novel architectures and training strategies to facilitate this integration.

\section{CONCLUSION}

    With ongoing advancements and research in community detection-based image segmentation, we can expect more accurate, efficient, and context-aware segmentation algorithms in the future. These algorithms will continue to find applications in a wide range of fields, contributing to advancements in medical diagnosis, visual understanding, and real-world applications of computer vision technology. In conclusion, community detection-based image segmentation techniques offer a powerful approach for accurately segmenting complex images. By utilizing the principles of community detection, these methods can effectively identify cohesive regions within an image, enabling improved object recognition, scene understanding, and image analysis. The potential for advancements in fields such as medical imaging, autonomous driving, and computer graphics is enormous as researchers continue to explore and refine these techniques. With further development and integration of deep learning, contextual information, and advanced evaluation metrics, community detection-based image segmentation has the potential to revolutionize the field of computer vision and open up opportunities for image analysis and interpretation.

\bibliographystyle{ieeetr}
\bibliography{citation} 

\end{document}